  \documentclass[journal]{IEEEtran}

  \usepackage[pdftex]{graphicx}
  
  \DeclareGraphicsExtensions{.pdf,.jpeg,.png}
  
  \usepackage{cite}
  \usepackage{amsmath}
  \usepackage[colorlinks,citecolor=blue]{hyperref}
  
  \usepackage{chngcntr} 
  
  \usepackage{graphicx}
  \usepackage{tikz}
  \usetikzlibrary{arrows.meta}  
  \usetikzlibrary{calc}
  
  \usepackage{booktabs}
  \usepackage{multirow}
  
  \usepackage{xcolor}
  \usepackage{listings}
  \usepackage{float}
  
  \newcommand{\code}[1]{\textit{#1}}
  
\begin{document}
  %
  \title{Video Compression for Spatiotemporal \\Earth System Data}
  
  %
  
  \author{Oscar J. Pellicer-Valero, Cesar Aybar, and Gustau Camps Valls,~\IEEEmembership{Fellow,~IEEE}
   \thanks{Image Processing Laboratory, Universitat de Val\`encia, Val\`encia, Spain (email: oscar.pellicer@uv.es).}
  \thanks{Manuscript submitted \today. The authors thank Julio Contreras for providing the SimpleS2 dataset to test the library. Work supported by the ERC under the ERC-SyG-2019 USMILE project `Understanding and Modelling the Earth System with Machine Learning' (grant agreement 855187), and the HORIZON projects THINKINGEARTH (grant agreement 101130544) and ELIAS (grant agreement 101120237).}%
  }
  
  \markboth{}
  {Oscar J. Pellicer-Valero \MakeLowercase{\code{et al.}}: Video compression of spatiotemporal Earth data}
  %
  
  \maketitle
  
  \begin{abstract}
  Large-scale Earth system datasets, from high-resolution remote sensing imagery to spatiotemporal climate model outputs, exhibit characteristics analogous to those of standard videos. Their inherent spatial, temporal, and spectral redundancies can thus be readily exploited by established video compression techniques. Here, we present \code{xarrayvideo}, a Python library for compressing multichannel spatiotemporal datasets by encoding them as videos. 
  Our approach achieves compression ratios of up to 250x while maintaining high fidelity by leveraging standard, well-optimized video codecs through \code{ffmpeg}. 
  We demonstrate the library's effectiveness on four real-world multichannel spatiotemporal datasets: DynamicEarthNet (very high resolution Planet images), DeepExtremeCubes (high resolution Sentinel-2 images), ERA5 (weather reanalysis data), and the SimpleS2 dataset (high resolution multichannel Sentinel-2 images), achieving approximate Peak Signal-to-Noise Ratios (PSNRs) of 55.86, 40.60, 46.58, and 43.23 dB at 0.1~bits per pixel per band (bpppb) and 65.91, 54.28, 62.90, and 55.04 dB at 1 bpppb. 
  Additionally, we are redistributing two of these datasets, DeepExtremeCubes (2.3~Tb) and DynamicEarthNet (525~Gb), in the machine-learning-ready and cloud-ready TACO format through HuggingFace at significantly reduced sizes (270~Gb and 8.5~Gb, respectively) without compromising quality (PSNR 55.77-56.65 and 60.15), using \code{xarrayvideo}. 
  No performance loss is observed when the compressed versions of these datasets are used in their respective deep learning-based downstream tasks (next step reflectance prediction and landcover segmentation, respectively). 
  In conclusion, \code{xarrayvideo} presents an efficient solution for handling the rapidly growing size of Earth observation datasets, making advanced compression techniques accessible and practical to the Earth science community. 
  The library is available for use at \url{https://github.com/IPL-UV/xarrayvideo}.
  \end{abstract}
  
  

  \section{Introduction}
  
  \begin{figure}[th]
      \centering
      \includegraphics[width=\linewidth]{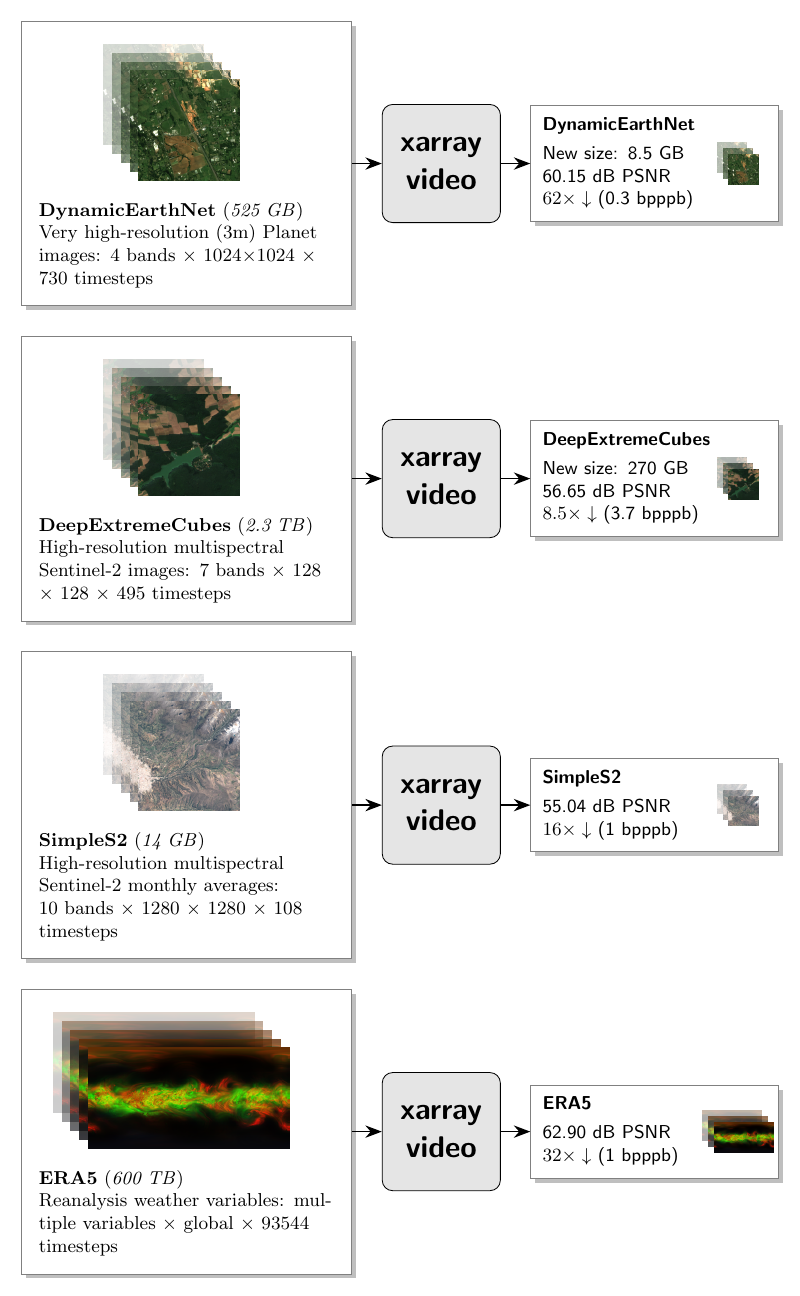}
      \caption{The \code{xarrayvideo} compression pipeline: Diverse Earth system datasets (DynamicEarthNet, DeepExtremeCubes, SimpleS2, and ERA5) are processed by \code{xarrayvideo}, leveraging standard video codecs, to produce significantly smaller compressed versions while maintaining high fidelity (PSNR). The compressed DynamicEarthNet and DeepExtremeCubes datasets are available on HuggingFace in the FAIR-compliant TACO format~\cite{aybar2025taco} at a $62\times$ and $8.5\times$ reduced size, respectively.}
      \label{fig:conceptual_compression}
  \end{figure}
  
  \IEEEPARstart{T}{he} field of Earth system sciences has experienced an unprecedented growth in recent years, as both public and private institutions begin to realize the potential across domains such as climate monitoring, urban planning, agriculture, and disaster management. To fuel these developments, Earth System Data (ESD) is also growing exponentially, further feeding discoveries and applications. ESD encompasses a variety of data sources, including: Earth Observation (EO) satellite missions, which are delivering imagery at ever-increasing spatial, temporal, and spectral resolutions \cite{UCSDatabase2023}; reanalysis data \cite{reichstein2019deep}, providing the most accurate snapshot of the planet's atmosphere over the last decades; and climate data, projecting those atmospheric variables hundreds of years into the future \cite{ipcc2022wg2}. Together, these provide invaluable insights into our planet's complex dynamics. Yet, for a large part of the scientific community, they remain beyond reach due to their sheer size.
  
  Indeed, the utility of ESD extends into many scientific and operational domains. In atmospheric sciences, ESD is fundamental for improving weather forecasting accuracy, understanding climate dynamics through modeling complex atmospheric, oceanic, and terrestrial interactions, and for producing detailed climate projections \cite{liu2020multimodal, Yang2013role, Gens2010remote}. Furthermore, satellite imagery and reanalysis data are routinely used to detect and monitor meteorological phenomena, such as extreme weather events like heatwaves, droughts, heavy rainfall, and floods, thereby supporting early warning systems for natural disasters \cite{taravat2015, oddo2019value} and climate studies \cite{Yang2013role}. ESD also plays a role in public health, for instance, in reconstructing historical air pollution levels to study their health impacts \cite{chen2019comparison}.
  
  In oceanography, ESD derived from sources like maritime radar and satellite altimetry helps characterize ocean wave parameters and monitor sea levels \cite{Gens2010remote}, which are crucial for navigation safety, coastal management, and understanding ocean-atmosphere exchanges. Remote sensing data also plays a pivotal role in terrestrial applications, providing a consistent global view of forests and urban areas. This includes monitoring land cover changes (e.g., deforestation, urban expansion), assessing natural resource availability, understanding the impacts of anthropogenic activities on ecosystems, and categorizing ecosystem types to inform biodiversity conservation and land management practices \cite{camps2011remote}. Furthermore, ESD supports decision-making in areas like food security and environmental health \cite{camps2021deep}, and is vital for monitoring progress towards Sustainable Development Goals (SDGs) \cite{persello2022deep}.
  
  Despite their profound relevance, these applications remain largely inaccessible to many due to the ever-increasing size of Earth system datasets, which only experts can realistically download and use. We are witnessing the rise of massive, multi-terabyte datasets needed for training contemporary Deep Learning (DL) and Foundation Models (FMs) \cite{Bommasani2021}, which brings forth significant technical hurdles, particularly in storage, transmission, and processing. Examples include BigEarthNet (100~Gb of Sentinel-1 \& 2 data) \cite{clasen2024reben}, SSL4EO (1~Tb of Sentinel-1 \& 2 data) \cite{wang2023ssl4eo}, and Major TOM (40~Tb of Sentinel-1 \& 2 data) \cite{francis2024major}. As the historical record of satellite missions extends and the benefits of leveraging the temporal dimension become clearer, spatiotemporal datasets capturing Earth's dynamics are also rapidly growing. Notable examples are DynamicEarthNet (525~Gb of PlanetFusion data) \cite{toker2022dynamicearthnet}, EarthNet21 (1~Tb of Sentinel-1 data) \cite{requena2021earthnet2021}, and DeepExtremeCubes (2.3~Tb of Sentinel-2 data) \cite{ji2024deepextremecubes}. This trend is set to accelerate, especially with the development of data-hungry FMs for EO. Concurrently, the demand for high-resolution climate and weather data products, such as the 600Tb ERA5 dataset \cite{hersbach2020era5}, intensifies, fueling next-generation models that integrate diverse ESD sources.
  
  Given the inherent redundancy in ESD across spatial, temporal, and spectral dimensions, data compression techniques, both lossless (no information) is lost) and lossy (some information is lost), are vital. Current approaches to multichannel image compression primarily include: general-purpose lossless algorithms (e.g., Zstandard), often coupled with domain-specific formats like NetCDF \cite{rew1990netcdf} and Zarr \cite{moore2023zarr}; image-specific standards like JPEG2000 and JPEG-XL \cite{alakuijala2019jpeg}, which can handle multiple channels but typically lack innate support for exploiting temporal correlations in image sequences; and neural compression methods such as autoencoders \cite{balle2018variational} and VQ-VAEs \cite{van2017neural}. While promising, neural methods are often computationally intensive, may not offer variable-rate compression \cite{i2024fixed}, and generally do not explicitly address time-series data beyond treating frames independently. Canonical methods like transform coding (e.g., DCT, wavelets) and predictive coding have been foundational but are mostly applied to still images or individual spatiotemporal fields rather than exploiting long-range temporal dependencies in sequences.
  
  A recent and notable example in the field of weather data is the Cra5 dataset by Han et al. \cite{han2024cra5}, a compressed 0.7~Tb version of the 226~Tb ERA5 weather dataset using a Variational Autoencoder Transformer; the authors showed no reduction in skill when training a forecasting model on the compressed data. Another recent proposal by Xiang et al. \cite{xiang2024remote} modifies the Ballé et al. \cite{balle2018variational} autoencoder architecture to process high and low frequencies separately for compression of Remote Sensing images, achieving competitive results for high compression ratios. Still, while these methods perform well for individual images, they all fail to exploit temporal correlations in time-series data. If we add the time component, compression of multichannel videos is currently vastly underexplored, with the only paper we found on the topic being the work by Das et al. \cite{das2021hyperspectral}, in which the difference between consecutive frames is first computed, each frame is then decomposed into a compact core tensor and factor matrices through sparse Tucker decomposition, and compression is achieved by truncating less significant components.
  
  A crucial topic that is rarely discussed on compression methods is implicit evaluation. Beyond explicit metrics such as compression ratio (often measured in bits per pixel or entropy) and reconstruction fidelity (e.g., PSNR, SSIM), it is also crucial to assess their implicit impact on the performance of downstream tasks, such as classification, retrieval, or physical parameter estimation \cite{garcia2019improved, Garcia17compress, Garcia-Vilchez2011253}. Garcia et al. \cite{garcia2010impact} go even one step further, observing that, for specific compression techniques and models, a higher compression ratio can lead to more accurate classification results, which can be explained in light of the spatial regularization and whitening that most compression techniques perform. We argue that a dataset compressed lossily at sufficiently high quality is completely interchangeable with the original data for any downstream application.
   
  In short, while DL-based methods show a lot of promise, they lack standardization, require extensive knowledge to apply to new datasets, are computationally expensive, and require specific hardware, limiting their practical adoption to general datasets and research scenarios. Most importantly, as was already discussed, all methods fail to take proper advantage of temporal correlations in time-series data. To address these issues, we propose a simpler yet effective solution: \code{xarrayvideo} (Figure~\ref{fig:conceptual_compression}), a Python library that leverages standard video codecs to compress multichannel spatiotemporal data efficiently. This is achieved using the \code{ffmpeg} library, which is widely available and accessible for all systems and contains well-optimized implementations of most video codecs. Thanks to the \code{xarray} library, \code{xarrayvideo} is also fully compatible with the existing geospatial data ecosystem, allowing simple integration with existing workflows and making it readily available for any dataset with minimal effort by the researcher. Interestingly, this approach had not been explored before, with the closest idea being to use video codecs to compress (non-temporal) hyperspectral data by treating the bands as the time dimension \cite{santos2011performance}. Perhaps this is due to the limitations of the approach, namely, most video codecs can only store up to 3 channels in 8-12~bit depths, restricting their direct application to EO datasets. However, this work demonstrates that standard video codecs can still be effectively applied to multispectral data compression.
  
  We introduce the following contributions: First, we present a new Python library, \code{xarrayvideo} (\url{https://github.com/IPL-UV/xarrayvideo}), for saving datasets (in \code{xarray} format) as videos using a variety of video codecs (through \code{ffmpeg}) and image codecs (through \code{GDAL}), with support for a variety of encoders (such as \code{libx265}, \code{hevc\_nvenc}, \code{vp9}, \code{ffv1}, etc. for video, and \code{JPEG2000} for image), both lossy and lossless, at different quality presets, different bit depths, etc. Second, we showcase the utility of the library through a set of compression benchmarks on four representative multichannel geospatial datasets, including an evaluation of the compressed data on several DL downstream tasks. Finally, we redistribute through HuggingFace, in the machine-learning-ready and cloud-ready TACO format~\cite{aybar2025taco}, a compressed version of the DeepExtremeCubes dataset (compressed from 2.3~Tb to 270~Gb at 55.765-56.653~dB PSNR) at \url{https://huggingface.co/datasets/isp-uv-es/DeepExtremeCubes-video} and the DynamicEarthNet dataset (compressed from 525~Gb to 8.5~Gb at 60.151~dB~PSNR) at \url{https://huggingface.co/datasets/isp-uv-es/DynamicEarthNet-video}, hence serving as illustrative examples, as well as providing to the community a much more accessible version of these datasets. 
  
  \section{Materials and methods}
  
  In this section, we review the main characteristics of the library provided, the data used in our intercomparisons (remote sensing, reanalysis, and climate spatiotemporal data), and discuss the metrics for evaluation.
   
  \subsection{Spatiotemporal data compression --- Library overview}
   
  \code{xarrayvideo} 
  includes two main functions: \code{xarray2video} and \code{video2xarray}, for converting \code{xarray}s to \code{xarrayvideo} (how we coined the resulting compressed format), and vice versa. This encoded format, \code{xarrayvideo}, in practice, is just a directory containing a series of standard video files (with some custom metainformation for decompression embedded), along with a single NetCDF zip-compressed xarray file, consistently named \code{x.nc}, which contains all the variables in the original \code{xarray} that were not explicitly compressed as videos. The main inputs to \code{video2xarray} are the \code{xarray} DataArray/Dataset and a configuration dictionary, describing how the mapping from the \code{xarray} variables to the resulting video files should be done.

  \begin{figure}[H]
  \begin{lstlisting}[language=Python, caption={Python dictionary with the mapping rules from the \code{xarray} variables to the resulting video files}, label=ls:config]
   mapping_rules = {
     'bands': ( # output video name
       ['R', 'G', 'B', 'NIR'], # input vars
       ('time', 'y', 'x'), # input tyxc coords
       0, # number of PCs (0 to disable PCA) 
       { 'c:v': 'libx265', # ffmpeg/GDAL config
         'preset': 'medium',
         'x265-params': 'qpmin=0:qpmax=0.01',
         'tune': 'psnr' }, 
       12, # output bit depth
     ), 
     'labels': ( # output video name
       'labels', # input vars
       ('time_month', 'y', 'x'), # input tyxc coords
       0, # number of PCs (0 to disable PCA) 
       { 'c:v': 'ffv1' }, # ffmpeg/GDAL config
       8, # output bit depth
     ),
   }
  \end{lstlisting}
\end{figure}
   
  Listing~\ref{ls:config} shows an example configuration that was used for generating the \code{xarrayvideo} version of the DynamicEarthNet dataset (see Section \ref{sec:dynamicearthnet}). For this particular configuration, four output files are produced for each \code{xarray} file: \code{bands\_0001.mkv}, \code{bands\_0002.mkv}, \code{labels\_0001.mkv}, and \code{x.nc}. For the \code{bands} videos, we are encoding four variables (R, G, B, and NIR), but the encoder that we are using, \code{libx265}, only supports videos of up to three bands, so the first three bands (R, G, B) are automatically split and saved in video \code{bands\_0001.mkv}, and the remaining bands (NIR) are repeated up to three times (e.g.: NIR, NIR, NIR) and encoded in video \code{bands\_0002.mkv}. Similarly, for the \code{labels} video, we encode a single variable (labels) into a video file, \code{labels\_0001.mkv}, using a different (lossless) encoder, \code{ffv1}.
   
  Internally, the library uses two backends for performing the encoding and decoding: \code{ffmpeg} and Python's \code{GDAL} library, though \code{GDAL} is mainly used for making comparisons between video encoding and frame-by-frame image encoding (the standard baseline encoding method). Out of the box, \code{xarrayvideo} supports video codecs \code{libx264}, \code{libx265}, \code{vp9}, \code{ffv1}, and \code{hevc\_nvenc} (though more can be trivially added, as long as they are supported by \code{ffmpeg}), as well as image codec \code{JP2OpenJPEG} (for \code{JPEG2000} compression). Depending on this availabe codec used, either \code{ffmpeg} or \code{GDAL} accept various configuration options, which can be provided through the mapping dictionary. For instance, in Listing~\ref{ls:config}, codec \code{libx265} is tuned to optimize PSNR, and the quantization parameter, which defines the final quality (lower is better), is forced to lie between 0 and 0.01. Note that this configuration dictionary could also be a list of dictionaries, providing a different configuration for each output video (for \code{bands\_0001.mkv}, \code{bands\_0002.mkv} in this case). Additionally, we can decide the number of bits to be used in the final compressed video: for the satellite data (video \code{bands}), we choose to use 12~bits, which is the maximum that \code{libx265} supports (though the original data is in 16~bits), while for the label we use 8~bits since the label is originally also encoded as an 8~bit integer.
   
  A final feature of the library is the ability to perform a Principal Components (PC) Analysis (PCA) transformation of the data before encoding it as a video. This is controlled by setting the third input in the mapping rules $>0$, defining the number of PCs to keep in the final compressed video(s). For instance, for video \code{bands} in Listing~\ref{ls:config}, if we set this number to 3, then only one output video would be produced, \code{bands\_0001.mkv} for all four R, G, B, NIR bands, containing their first 3 PCs, and potentially providing better compression ratios.
   
  \subsection{Benchmark datasets}
   
  We evaluated \code{xarrayvideo} on four real-world datasets involving spatiotemporal data modalities: DynamicEarthNet (very high-resolution Planet image sequences)~\cite{toker2022dynamicearthnet}, DeepExtremeCubes (high resolution multispectral Sentinel-2 image sequences)~\cite{ji2024deepextremecubes}, ERA5 (reanalysis data sequences of several weather variables)~\cite{hersbach2020era5}, and SimpleS2 (a custom Sentinel-2 dataset explicitly created for testing the library), hence providing a robust set of test cases to evaluate the effectiveness of video-based compression.
   
  \subsubsection{DynamicEarthNet}\label{sec:dynamicearthnet}
  A collection of 75 cubes sampled from 75 selected areas of interest distributed over the globe with imagery from Planet Labs, each cube with dimensions time (730) $\times$ latitude (1024) $\times$ longitude (1024) $\times$ bands (four bands: B04, B03, B02, B8A) at daily 3 m resolution from January 2018 to December 2019. Each cube is paired with pixel-wise monthly semantic segmentation labels of seven land use and land cover classes~\cite{toker2022dynamicearthnet}. Five cubes were randomly selected for benchmarking, and the four spectral bands were compressed.
   
  \subsubsection{DeepExtremeCubes}
  Around 40.000 Sentinel-2 data cubes were sampled from a wide range of locations and land covers, with a strong focus on areas impacted by meteorological extreme events~\cite{ji2024deepextremecubes}. Each data cube is of dimensions time (495) $\times$ latitude (128) $\times$ longitude (128) $\times$ bands (seven bands: B04, B03, B02, B8A, B05, B06, B07) at five-daily 30 m resolution. In addition to the spectral bands, each cube contains a variety of additional dynamic masks, such as a cloud mask and a scene classification layer (SCL) with the same timesteps and resolution as the spectral data, as well as a variety of static data (vegetation landcover and digital elevation map), and temporal data (ERA5-derived variables, such as temperature, pressure, etc.). For benchmarking, ten randomly picked cubes were selected, and video compression was tested on three sets of data: the RGB spectral bands (B04, B03, B02, which were lossily encoded), three infrared bands (B11, B08, B09, which also were lossily encoded), and three dynamic masks (cloud mask, SCL, and a validity mask), which were losslessly encoded.
  
  \subsubsection{SimpleS2}
  Four Sentinel-2 data cubes from four locations encompassing a variety of land covers were specifically picked to assess the effectiveness of applying PCA before compression. Each cube is of dimensions time (108) $\times$ latitude (1280) $\times$ longitude (1280) $\times$ bands (ten bands: B02, B03, B04, B05, B06, B07, B08, B8A, B11, B12), each timestep being a monthly average (from 10-2015 to 09-2024), and each pixel being at 10 m resolution. The cubes had been processed to exclude clouds, and any remaining gaps were linearly interpolated between adjacent months. All the data in this dataset was used for benchmarking. 
  This dataset is available at \url{https://huggingface.co/datasets/isp-uv-es/SimpleS2}.

  \subsubsection{ERA5}
  The European Centre for Medium-Range Weather Forecasts (ECMWF) reanalysis data encompasses a detailed record of the global atmosphere, land surface, and ocean waves from 1950 onwards~\cite{hersbach2020era5}. In practice, we used a subset of the data provided by WeatherBench2~\cite{rasp2024weatherbench} consisting of a single cube (resampled to 6h intervals) of dimensions time (93544) $\times$ latitude (721) $\times$ longitude (1440) $\times$ level (13) containing a variety of meteorological variables (such as pressure, temperature, wind speed, etc.), encompassing the whole Earth from 1959 to 2023 (included). For benchmarking, 284 timesteps of the variables' magnitude of wind, temperature, and precipitation were considered at the 13 different pressure levels. 
   
  \subsection{Evaluation metrics}
  
  The compression performance was evaluated using both size and quality metrics. For different data sizes, we measured compression efficiency using bits per pixel per band (bpppb). To put this into perspective, an uncompressed image using standard 32-bit floating-point numbers requires 32 bpppb, since storing the value of one pixel (for one band) takes up 32 bits of storage space; hence, lower bpppb values indicate better compression. Conversely, compressing that same image at 1 bpppb would imply a 32x compression rate. For quality assessment, we employed three complementary metrics: 
  \begin{itemize}
  \item Peak Signal-to-Noise Ratio (PSNR) \cite{jahne2005digital}, which measures the ratio between maximum possible signal power and noise power on a logarithmic scale (higher is better). In this work, PSNR is computed with respect to the maximum intensity value present in the original image sequence for each channel, as this reflects the actual dynamic range of the data. It is important to note that some literature may calculate PSNR with respect to a fixed maximum (e.g., \(2^B-1\) for B-bit data), which can lead to misleadingly high PSNR values, as data does not typically span the full bit-depth range. 
  \item Structural Similarity Index (SSIM) \cite{wang2004image}, which evaluates perceived image quality by analyzing structural information (higher is better)
  \item Spectral Angle (SA) \cite{kruse1993spectral}, which measures the angle between original and reconstructed spectral signatures (lower is better).
  \end{itemize}
  To generate the lines in figures~\ref{fig:dynamicearthnet} to~\ref{fig:era5}, we used Lowess \cite{moran1984locally} non-parametric interpolation over several cubes (between one and ten, depending on the dataset) and over six different quality settings (from ``Best" to ``Very low"), the configurations for which can be found in Supplementary Table~1. We also measured compression and decompression times to evaluate computational efficiency, which can be found in tables~\ref{tab:dynamicearthnet} to~\ref{tab:era5}. Together, these metrics comprehensively assess the compression performance across different aspects, such as size reduction, visual quality preservation, spectral fidelity, and processing overhead. 
   
  \section{Experimental Results}
  
  We present the experimental results obtained by applying \code{xarrayvideo} to the benchmark datasets, from remote sensing high-resolution multispectral images to spatiotemporal reanalysis data, demonstrating the versatility and effectiveness of our approach for Earth system data. We also evaluate various compression configurations, comparing different codecs, bit depths, and the impact of using PCA to reduce the number of bands to be compressed.
   
  \subsection{Very high-resolution satellite images --- DynamicEarthNet}
   
  \begin{figure*}[!t]
       \centering
   
       \begin{tikzpicture}
           \def\imagewidth{0.12}  
           \def\spacing{0.04}     
           \def\spacingmultiplier{1.04}     
           \def\cropwidth{0.125} 
           \def\cropheight{0.875} 
           
           \node[anchor=south west,inner sep=0] (first_image) at (-\imagewidth\textwidth,0) {
               \includegraphics[width=\imagewidth\textwidth]{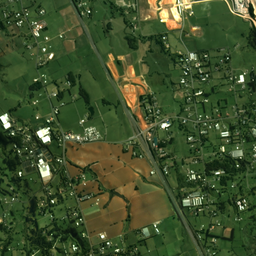}
           };
           \node[below, font=\footnotesize] at (first_image.south) {Uncropped};
           
           \node[anchor=south west,inner sep=0] (second_image) at (\spacing,0) {
               \includegraphics[width=\imagewidth\textwidth]{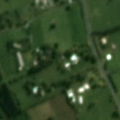}
           };
           \node[below, font=\footnotesize, align=center] at (second_image.south) {Original \\ 16~bpppb \\ $\infty$~dB};
           
           \begin{scope}[x={(first_image.south east)},y={(first_image.north west)}]
               \draw[red, thick] (0,1) rectangle +(-\cropwidth,-\cropwidth);
           \end{scope}
           
           \begin{scope}[x={(second_image.south east)},y={(second_image.north west)}]
               \draw[red, thick] (0,1) rectangle (1,0);
           \end{scope}
           
           \foreach \idx/\quality/\bpppb/\psnr [count=\i from 0] in {
               Best/Best/2.865/75.291,
               Very high/Very\_high/0.197/59.217,
               High/High/0.120/57.918,
               Medium/Medium/0.059/54.379,
               Low/Low/0.018/49.260,
               Very low/Very\_low/0.004/43.071
           } {
               \pgfmathsetmacro{\xpos}{(\imagewidth*\spacingmultiplier)*(\i+1)}
               \pgfmathsetmacro{\labelpos}{\xpos+0.065}
               \node[anchor=south west,inner sep=0] at (\xpos\textwidth,0) {
                   \includegraphics[width=\imagewidth\textwidth]{dynamicearthnet_img_libx265_12_8077_5007_13_bands_\quality_60_comp_crop_topleft.png}
               };
               \node[below, font=\footnotesize, align=center] at (\labelpos\textwidth,0) {\quality \\ \bpppb~bpppb \\ \psnr~dB};
           }
       \end{tikzpicture}
   
       \begin{minipage}{0.33\textwidth}
           \includegraphics[width=\textwidth]{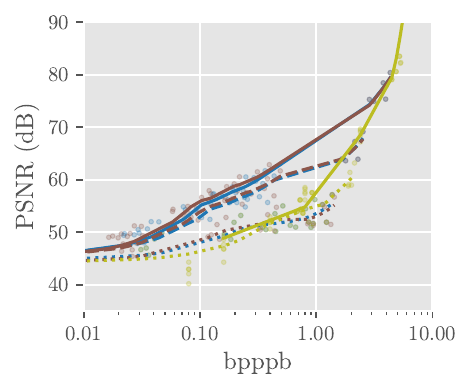}
       \end{minipage}%
       \begin{minipage}{0.33\textwidth}
           \includegraphics[width=\textwidth]{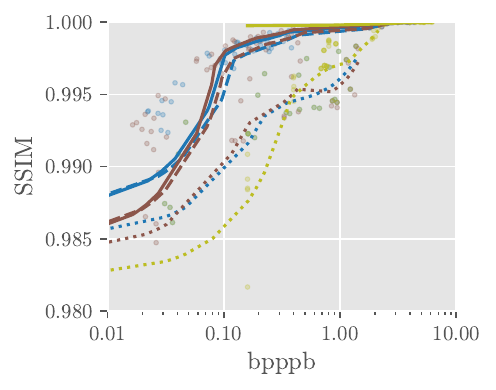}
       \end{minipage}%
       \begin{minipage}{0.33\textwidth}
           \includegraphics[width=\textwidth]{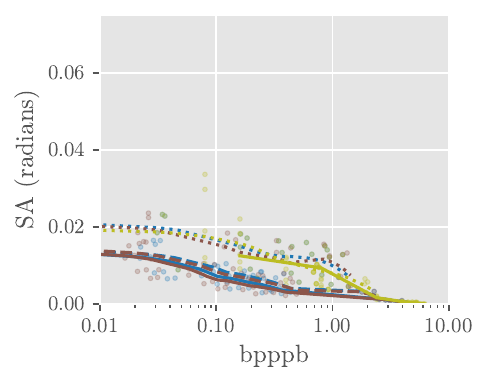}
       \end{minipage}
       \vspace{2mm}
       \includegraphics[width=0.6\textwidth]{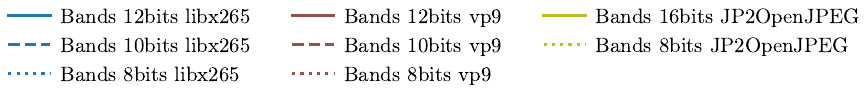}
       \caption{Top: Sample image from the DynamicEarthNet dataset showing RGB bands (original is a 16-bit integer in range 0-65535) after scaling by a factor of 1/2000 and clipping between 0 and 1 for better visualization; from left to right: original, crop (region within the red square in the original image), crops of compressed versions using \code{x265} 12 bits at different quality settings. Bottom: Compression performance metrics for the DynamicEarthNet dataset across three quality metrics: (a) Peak Signal-to-Noise Ratio (PSNR), (b) Structural Similarity Index (SSIM), and (c) Spectral Angle (SA) plotted against the achieved compression rate in bits per pixel per band (bpppb), for a variety of codecs (x265, vp9 and JPEG2000), at different bit depths (8, 10, 12, or 16). Higher values are better for PSNR and SSIM, while lower values are better for SA.}
       \label{fig:dynamicearthnet}
  \end{figure*}

\begin{table}[h!]
    \centering
    \caption{Comparison of different compression configurations for the DynamicEarthNet dataset. bpppb: bits per pixel per band, PSNR: Peak Signal-to-Noise Ratio, $t_{c}$: compression time (in seconds), $t_{d}$: decompression time (in seconds)}
    \label{tab:dynamicearthnet}
    \begin{tabular}{llrrrrr}
\toprule
& & \multicolumn{4}{c}{Bands} \\
Test & Bits & Quality & bpppb & PSNR & $t_{c}$ & $t_{d}$\\
\midrule
\multirow{6}{*}{\shortstack{JP2\\Open\\JPEG}} & \multirow{6}{*}{16} & Best & 5.649 & $\infty$ & 129.738 & 76.477 \\
& & Very high & 4.774 & 80.687 & 178.166 & 72.180 \\
& & High & 4.764 & 80.658 & 179.164 & 71.719 \\
& & Medium & 2.400 & 68.764 & 166.360 & 54.374 \\
& & Low & 0.800 & 57.581 & 151.416 & 39.944 \\
& & Very low & 0.160 & 46.831 & 133.503 & 33.949 \\
\midrule
\multirow{6}{*}{\shortstack{lib\\x265}} & \multirow{6}{*}{12} & Best & 3.689 & 76.882 & 276.179 & 22.611 \\
& & Very high & 0.295 & 60.151 & 105.798 & 11.351 \\
& & High & 0.164 & 58.179 & 89.026 & 10.288 \\
& & Medium & 0.076 & 54.556 & 70.895 & 8.997 \\
& & Low & 0.021 & 49.392 & 49.377 & 8.264 \\
& & Very low & 0.004 & 43.249 & 39.236 & 8.484 \\
\midrule
\multirow{6}{*}{vp9} & \multirow{6}{*}{12} & Best & 2.880 & 76.882 & 790.632 & 18.788 \\
& & Very high & 2.880 & 76.882 & 790.638 & 18.732 \\
& & High & 0.180 & 56.686 & 961.768 & 8.862 \\
& & Medium & 0.048 & 51.355 & 916.003 & 8.073 \\
& & Low & 0.024 & 48.755 & 858.408 & 8.180 \\
& & Very low & 0.012 & 46.145 & 841.540 & 8.199 \\
\bottomrule
\end{tabular}
\end{table}

  The results of encoding DynamicEarthNet with \code{xarrayvideo} can be seen in Figure~\ref{fig:dynamicearthnet}, as well as 
  Table~\ref{tab:dynamicearthnet}. Figure~\ref{fig:dynamicearthnet} shows a sample image compressed at different levels, as well as quality metrics PSNR, SSIM, and SA against bpppb for a variety of codec and bit depth combinations, providing some interesting insights: Looking at the bottom plots, video codecs \code{x265} and \code{vp9} provide a very similar compression performance, dominating image codec \code{JPEG2000} on all metrics (except perhaps SSIM), for all bpppb values except for the higher end (4 to 5 bpppb). Also, higher bit depths (12~bits for video and 16~bits for image) consistently outperform lower bit depths (e.g., 8~bits), which is expected since the native source was at 16~bits, especially as the bpppb increase. This is likely why 16~bit \code{JPEG2000} outperforms video codecs at the highest bpppb. Looking at the top plots, we can see crops of compressed versions of a sample image (to the left) using \code{x265} 12 bits at decreasing quality settings, hence providing intuition for the metrics in the plots below (see 
  table \ref{tab:codec_configs} for the employed codec configuration at each quality level). We observe that for the DynamicEarthNet dataset, the quality is extremely high even at the lowest bpppb values, with noticeable artifacts only starting to appear at the very low settings (at 43 dB PSNR and a 4000x compression ratio).

  From 
  Table~\ref{tab:dynamicearthnet}, the median compression times amounts to 79.96~s, 849.97~s, and 158.89~s for codecs \code{x265}, \code{vp9} and \code{JPEG2000}, respectivley. Similarly, the median decompression times are 9.64~s, 8.53~s, and 63.05~s for those same codecs. Since \code{x265} is the quickest in compression (by a factor $>10$, when compared with \code{vp9}), similarly quick as \code{vp9} in decompression, and achieves identical quality to \code{vp9}, we consider it the best overall choice for this dataset. 
   
  \begin{table}[h!]
       \caption{Downstream task performance for the original and compressed versions (8.5~Gb at 60~dB PSNR, 2.1~Gb at 54~dB PSNR) of the DynamicEarthNet dataset. To generate this table, the pretrained models from \cite{toker2022dynamicearthnet} with weekly sampling were applied without modification to the original and compressed data. All metrics are reported as percentages; a higher percentage is better. mIoU: mean Intersection Over Union, Acc.: Accuracy.}
       \centering
       \setlength{\tabcolsep}{3pt}  
       \begin{tabular}{llcccc}
       \toprule
       \multirow{2}{*}{Model} & \multirow{2}{*}{Dataset} & \multicolumn{2}{c}{Validation} & \multicolumn{2}{c}{Test} \\
       \cmidrule(lr){3-4} \cmidrule(lr){5-6}
       & & mIoU & Acc & mIoU & Acc. \\
       \midrule
       \multirow{3}{*}{U-TAE} & Original (525~Gb) & 35.710 & 73.843 & 39.666 & 72.169 \\
       & Video 60dB (8.5~Gb) & 35.715 & 73.848 & 39.656 & 72.169 \\
       & Video 54dB (2.1~Gb) & 35.709 & 73.848 & 39.642 & 72.161 \\
       \midrule
       \multirow{3}{*}{U-ConvLSTM} & Original (525~Gb) & 34.331 & 73.203 & 39.118 & 71.457 \\
       & Video 60dB (8.5~Gb) & 34.325 & 73.192 & 39.108 & 71.449 \\
       & Video 54dB (2.1~Gb) & 34.313 & 73.186 & 39.091 & 71.428 \\
       \midrule
       \multirow{3}{*}{3D-Unet} & Original (525~Gb) & 35.459 & 73.798 & 37.182 & 68.393 \\
       & Video 60dB (8.5~Gb) & 35.469 & 73.779 & 37.187 & 68.391 \\
       & Video 54dB (2.1~Gb) & 35.479 & 73.770 & 37.205 & 68.404 \\
       \bottomrule
       \end{tabular}
       \label{tab:downstream_dynamicearthnet}
   \end{table}
  
  To further demonstrate the usefulness of the compressed dataset beyond general quality metrics such as PSNR, we evaluated the validation and test performance of the pretrained DL models from the original DynamicEarthNet publication~\cite{toker2022dynamicearthnet} using two \code{xarrayvideo} versions of the original dataset: A 2.1~Gb version at 54~dB PSNR and an 8.5~Gb at 60~dB PSNR. The downstream task was semantic segmentation of the satellite images over several landcover classes (impervious surface, agriculture, forest, wetlands, soil, water, and snow \& ice). The pretrained models from \cite{toker2022dynamicearthnet} were applied without modifying the original and the compressed data. Only weekly sampling results are reported here for simplicity, since it was found that weekly sampling is optimal. As shown in Table~\ref{tab:downstream_dynamicearthnet}, the compressed datasets achieve identical scores to the original one, with the 54~dB PSNR dataset being sometimes even better at just 2.1~Gb (i.e., a compression factor of 249x). The 8.5~Gb \code{xarrayvideo} version of this dataset (i.e., a compression factor of 62x) has been made openly available at \url{https://huggingface.co/datasets/isp-uv-es/DynamicEarthNet-video}. Note that we did not include the quality assurance metadata in this compressed dataset, and the included labels are the corrected version 2 \cite{toker_2024_dynamicearthnet}.

  \subsection{DeepExtremeCubes --- High resolution satellite images}  
  \begin{table*}[ht]
    \centering
    \caption{Comparison of different compression configurations for the DeepExtremeCubes dataset. bpppb: bits per pixel per band, PSNR: Peak Signal-to-Noise Ratio, $t_{c}$: compression time (in seconds), $t_{d}$: decompression time (in seconds)}
    \label{tab:deepextremecubes}
    \begin{tabular}{llrrrrrrrrr}
\toprule
& & \multicolumn{4}{c}{Ir3} & \multicolumn{4}{c}{Rgb} \\
Test & Bits & Quality & bpppb & PSNR & $t_{c}$ & $t_{d}$ & bpppb & PSNR & $t_{c}$ & $t_{d}$\\
\midrule
\multirow{6}{*}{\shortstack{JP2\\Open\\JPEG}} & \multirow{6}{*}{16} & Best & 11.857 & 107.393 & 23.888 & 9.315 & 11.121 & 107.543 & 23.267 & 9.164 \\
& & Very high & 11.231 & 100.034 & 24.617 & 9.345 & 10.470 & 99.997 & 23.918 & 9.145 \\
& & High & 5.576 & 67.805 & 22.697 & 8.806 & 5.576 & 71.689 & 21.264 & 9.128 \\
& & Medium & 2.380 & 48.608 & 21.946 & 8.353 & 2.382 & 52.036 & 19.385 & 8.244 \\
& & Low & 0.787 & 35.935 & 21.515 & 8.379 & 0.785 & 38.727 & 19.630 & 8.200 \\
& & Very low & 0.159 & 25.555 & 19.842 & 7.820 & 0.157 & 26.899 & 18.967 & 7.829 \\
\midrule
\multirow{6}{*}{\shortstack{lib\\x265}} & \multirow{6}{*}{12} & Best & 4.939 & 75.138 & 15.398 & 0.534 & 4.379 & 75.161 & 15.356 & 0.523 \\
& & Very high & 1.408 & 55.765 & 11.347 & 0.358 & 1.108 & 56.653 & 9.621 & 0.330 \\
& & High & 0.621 & 49.545 & 7.687 & 0.289 & 0.509 & 50.915 & 6.556 & 0.271 \\
& & Medium & 0.369 & 46.505 & 6.056 & 0.232 & 0.314 & 47.866 & 5.324 & 0.223 \\
& & Low & 0.126 & 41.141 & 3.862 & 0.174 & 0.120 & 42.331 & 3.545 & 0.171 \\
& & Very low & 0.033 & 35.306 & 2.545 & 0.158 & 0.033 & 36.022 & 2.359 & 0.160 \\
\midrule
\multirow{6}{*}{vp9} & \multirow{6}{*}{12} & Best & 4.392 & 75.138 & 20.921 & 0.481 & 3.936 & 75.161 & 20.096 & 0.511 \\
& & Very high & 4.392 & 75.138 & 20.916 & 0.502 & 3.936 & 75.161 & 20.081 & 0.447 \\
& & High & 0.821 & 51.271 & 49.598 & 0.229 & 0.650 & 52.478 & 40.775 & 0.220 \\
& & Medium & 0.292 & 44.882 & 38.800 & 0.194 & 0.250 & 46.241 & 33.115 & 0.179 \\
& & Low & 0.163 & 42.036 & 33.352 & 0.168 & 0.148 & 43.253 & 29.241 & 0.164 \\
& & Very low & 0.089 & 39.301 & 29.241 & 0.157 & 0.084 & 40.282 & 26.161 & 0.148 \\
\midrule
& & \multicolumn{4}{c}{masks}\\
\midrule
\multirow{1}{*}{ffv1} & \multirow{1}{*}{8} & Best & 0.235 & $\infty$ & 0.279 & 0.136 \\
\bottomrule
\end{tabular}
\end{table*}

  \begin{figure*}[!t]
   
       \centering
       \begin{tikzpicture}
           \def\imagewidth{0.137}  
           \def\spacing{0.04}     
           \def\spacingmultiplier{1.04}     
           \def\cropwidth{1} 
           \def\cropheight{1} 
           
           \node[anchor=south west,inner sep=0] (first_image) at (-\imagewidth\textwidth,0) {
               \includegraphics[width=\imagewidth\textwidth]{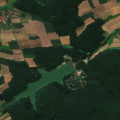}
           };
           \node[below, font=\footnotesize, align=center] at (first_image.south) {Original \\ 32~bpppb \\ $\infty$~dB};
   
           \begin{scope}[x={(first_image.south east)},y={(first_image.north west)}]
               \draw[red, thick] (0,1) rectangle (1,0);
           \end{scope}
           
           \foreach \idx/\quality/\bpppb/\psnr [count=\i from 0] in {
               Best/Best/5.371/75.101,
               Very high/Very\_high/1.338/56.350,
               High/High/0.612/50.595,
               Medium/Medium/0.339/47.340,
               Low/Low/0.113/42.223,
               Very low/Very\_low/0.030/36.678
           } {
               \pgfmathsetmacro{\xpos}{(\imagewidth*\spacingmultiplier)*(\i)+\spacing*\imagewidth}
               \pgfmathsetmacro{\labelpos}{\xpos+0.065}
               \node[anchor=south west,inner sep=0] at (\xpos\textwidth,0) {
                   \includegraphics[width=\imagewidth\textwidth]{deepextremes_img_libx265_12_10.38_50.15_rgb_\quality_120_comp_crop_center.png}
               };
               \node[below, font=\footnotesize, align=center] at (\labelpos\textwidth,0) {\quality \\ \bpppb~bpppb \\ \psnr~dB};
           }
       \end{tikzpicture}
   
       \begin{minipage}{0.33\textwidth}
           \includegraphics[width=\textwidth]{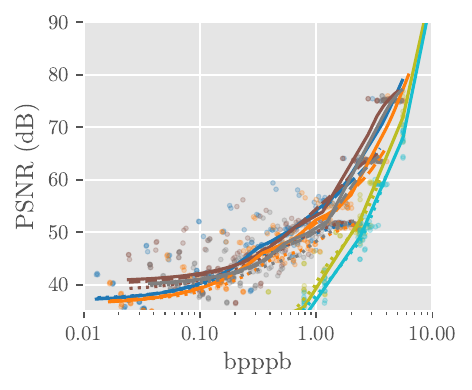}
       \end{minipage}%
       \begin{minipage}{0.33\textwidth}
           \includegraphics[width=\textwidth]{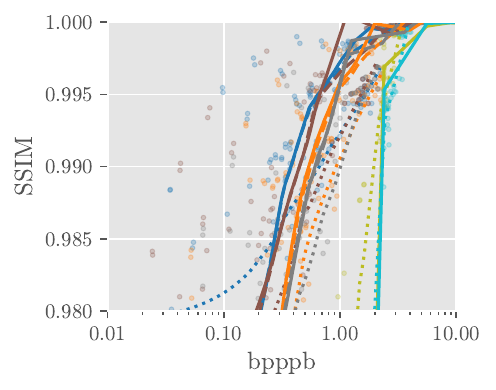}
       \end{minipage}%
       \begin{minipage}{0.33\textwidth}
           \includegraphics[width=\textwidth]{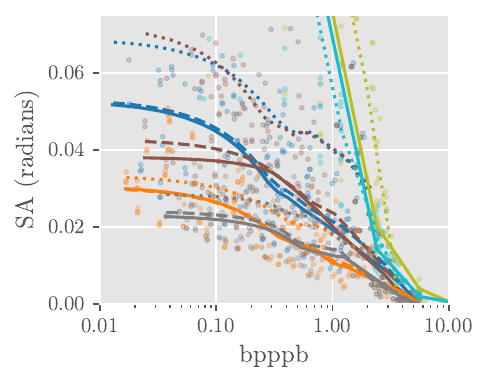}
       \end{minipage}
       \vspace{2mm}
       \includegraphics[width=\textwidth]{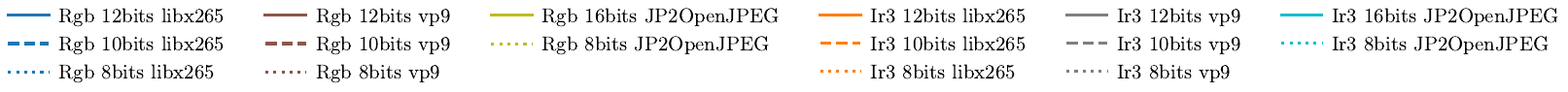}
       \caption{Top: Sample image from the DeepExtremeCubes dataset showing RGB bands after scaling by a factor of 4 and clipping between 0 and 1; from left to right: original, compressed versions using \code{x265} 12 bits at different quality settings. Bottom:  Compression performance metrics for the DeepExtremeCubes dataset across three quality metrics: (a) Peak Signal-to-Noise Ratio (PSNR), (b) Structural Similarity Index (SSIM), and (c) Spectral Angle (SA) plotted against the achieved compression rate in bits per pixel per band (bpppb), for two sets of three bands (RGB or IR3), for a variety of codecs (x265, vp9 and JPEG2000), at different bit depths (8, 10, 12, or 16). Higher values are better for PSNR and SSIM, while lower values are better for SA.}
       \label{fig:deepextremes}
  \end{figure*}
   
  For the DeepExtremeCubes dataset, Figure \ref{fig:deepextremes} represents a sample compressed image and several quality metrics against bpppb for several codecs and bit combinations. Overall, the compression quality is lower than that of the previous dataset at the same bpppb. This is likely due to a variety of factors making compression more difficult: first, clouds (more than 50\% of the images contain at least 50\% of cloud coverage~\cite{pellicer2024explainable}) and missing data add large intensity changes between consecutive timesteps; second, the imperfect registration between adjacent slices of Sentinel-2 images; and third, the smaller size of the images ($128\times 128$), disabling some encoding techniques. For this test, two sets of channels are compressed separately: RGB (containing standard red, green and blue bands), and IR3 (containing three infrared bands). Since video codecs are designed for RGB video compression, RGB is expected to perform better than IR3, yet this difference is very small, and, for SA, IR3 dominates. This result vouches for standard video codecs for compressing any bands, not just RGB. Inspecting the sample image at the top, we can visually identify artifacts at medium-quality settings (around 47 dB PSNR and a compression ratio of 94x), with extreme blurring appearing at the lowest-quality settings.
   
  In addition to the compression results for RGB and IR3, 
  Table~\ref{tab:deepextremecubes} shows the results of using codec \code{ffv1} for compressing three 8~bit masks (SCL, cloud mask, and a validity mask), achieving 0.235~bpppb (34x) for lossless compression, at compression and decompression times of 0.28s and 0.14s, respectively. For RGB and IR3, the decompression time of \code{libx265} at 12~bits is around 0.34~s. The rest of conclusions gathered from DynamicEarthNet (\code{x265} $\approx$ \code{vp9} $\gg$ \code{JPEG2000}, and 12~bits $> 10$~bits $\gg$ 8~bits) still apply for this data.
  
  Similar to the DynamicEarthNet dataset, we tested the downstream performance of the model proposed by Pellicer et al. \cite{pellicer2024explainable} by evaluating it on two \code{xarrayvideo} versions of the dataset. The task consisted in predicting the reflectances for bands B02, B03, B04, and B8A at the next timestep using a convolutional LSTM architecture. As Table~\ref{tab:downstream_deepextremecubes} shows, the performance for the 56~dB PSNR version of the dataset is identical to that of the original data, but at a compression factor of 8.5x; for the yet smaller 47~dB dataset (38x compression), we observe a small yet non-negligible impact in performance. The 270~Gb \code{xarrayvideo} version of this dataset has been made available at \url{https://huggingface.co/datasets/isp-uv-es/DeepExtremeCubes-video}. Note that channel B07, given its redundancy with channels B06 and B8A is not included from the original dataset to improve the compression ratio by requiring only two video files to encode all bands
  ; also, missing data has been filled forward (holding the last valid value), and an additional ``invalid" mask has been added since compressed data cannot hold missing values.
   
  \begin{table}[ht]
       \centering
       \setlength{\tabcolsep}{3pt}  
       \caption{Downstream task performance for the original and compressed versions (270~Gb at 56~dB PSNR, and 85~Gb at 47~dB PSNR) of the DeepExtremeCubes dataset. To generate this table, the pretrained model from \cite{pellicer2024explainable} was applied without modification to the original and compressed data. $R^2$: coefficient of determination (higher is better), $\mathcal{L}_1$: mean absolute error (lower is better), MSE: mean squared error (lower is better).}
       \label{tab:downstream_deepextremecubes}
       \begin{tabular}{lcccccc}
       \toprule
       & \multicolumn{3}{c}{Validation} & \multicolumn{3}{c}{Test} \\
       \cmidrule(lr){2-4} \cmidrule(lr){5-7}
       Dataset & $R^2$ & $\mathcal{L}_1$ & MSE & $R^2$ & $\mathcal{L}_1$ & MSE \\
       \midrule
       Original (3,2~Tb) & 0.883 & 0.041 & 0.079 & 0.906 & 0.036 & 0.070 \\
       Video 56dB (270~Gb) & 0.883 & 0.042 & 0.079 & 0.905 & 0.037 & 0.070 \\
       Video 47dB (85~Gb) & 0.876 & 0.045 & 0.082 & 0.899 & 0.040 & 0.073 \\
       \bottomrule
       \end{tabular}
  \end{table}
   
  \subsection{SimpleS2 --- High resolution multispectral images}

  \begin{table}[!ht]
      \centering
      \caption{Comparison of different compression configurations for the SimpleS2 dataset. For the test ``libx265 (PCA - 9 bands)", PCA was applied over the ten bands before compression, and the least important principal component (PC) was discarded so that all bands could fit in just three videos of three channels each. For the test ``libx265 (PCA - all bands)'', PCA was applied over the ten bands before compression, and all ten PCs were compressed differently according to their importance (the lower the importance, the higher the compression). bpppb: bits per pixel per band, PSNR: Peak Signal-to-Noise Ratio, $t_{c}$: compression time (in seconds), $t_{d}$: decompression time (in seconds)}
      \label{tab:custom}
      \begin{tabular}{llrrrrr}
  \toprule
  & & \multicolumn{4}{c}{All} \\
  Test & Bits & Quality & bpppb & PSNR & $t_{c}$ & $t_{d}$\\
  \midrule
  \multirow{6}{*}{\shortstack{JP2\\Open\\JPEG}} & \multirow{6}{*}{16} & Best & 8.288 & $\infty$ & 64.119 & 35.193 \\
  & & Very high & 8.250 & 94.043 & 108.898 & 35.088 \\
  & & High & 5.600 & 78.215 & 102.189 & 29.879 \\
  & & Medium & 2.400 & 59.176 & 89.052 & 21.175 \\
  & & Low & 0.800 & 47.412 & 81.437 & 16.245 \\
  & & Very low & 0.160 & 38.680 & 70.603 & 13.301 \\
  \midrule
  \multirow{6}{*}{\shortstack{lib\\x265}} & \multirow{6}{*}{12} & Best & 5.980 & 74.074 & 184.076 & 15.430 \\
  & & Very high & 1.176 & 56.070 & 102.539 & 10.191 \\
  & & High & 0.528 & 51.070 & 79.343 & 8.595 \\
  & & Medium & 0.246 & 47.182 & 63.406 & 7.620 \\
  & & Low & 0.068 & 42.063 & 46.620 & 6.353 \\
  & & Very low & 0.012 & 36.705 & 32.684 & 6.349 \\
  \midrule
  \multirow{6}{*}{\shortstack{lib\\x265\\-\\PCA\\9\\bands}} & \multirow{6}{*}{12} & Best & 4.472 & 59.147 & 143.478 & 36.164 \\
  & & Very high & 0.985 & 50.628 & 88.737 & 32.143 \\
  & & High & 0.511 & 47.251 & 67.822 & 30.885 \\
  & & Medium & 0.203 & 43.537 & 48.234 & 30.457 \\
  & & Low & 0.041 & 38.681 & 31.729 & 28.707 \\
  & & Very low & 0.005 & 33.667 & 21.944 & 28.973 \\
  \midrule
  \multirow{6}{*}{\shortstack{lib\\x265\\-\\PCA\\all\\bands}} & \multirow{6}{*}{12} & Best & 3.627 & 58.566 & 155.922 & 13.016 \\
  & & Very high & 0.644 & 52.173 & 81.791 & 8.689 \\
  & & High & 0.338 & 48.565 & 66.601 & 7.649 \\
  & & Medium & 0.146 & 44.407 & 52.258 & 6.774 \\
  & & Low & 0.042 & 39.545 & 38.753 & 6.345 \\
  & & Very low & 0.008 & 33.967 & 28.667 & 5.995 \\
  \midrule
  \multirow{6}{*}{vp9} & \multirow{6}{*}{12} & Best & 4.985 & 74.074 & 755.703 & 16.818 \\
  & & Very high & 4.985 & 74.074 & 751.353 & 16.729 \\
  & & High & 0.689 & 52.203 & 1108.209 & 8.160 \\
  & & Medium & 0.208 & 46.099 & 922.659 & 7.113 \\
  & & Low & 0.106 & 43.347 & 786.154 & 6.728 \\
  & & Very low & 0.050 & 40.809 & 743.289 & 6.325 \\
  \bottomrule
      \end{tabular}
  \end{table}
  
  \begin{figure*}[!t]
   
       \begin{tikzpicture}
           \def\imagewidth{0.12}  
           \def\spacing{0.04}     
           \def\spacingmultiplier{1.04}     
           \def\cropwidth{0.1067} 
           \def\cropheight{0.8933} 
           
           \node[anchor=south west,inner sep=0] (first_image) at (-\imagewidth\textwidth,0) {
               \includegraphics[width=\imagewidth\textwidth]{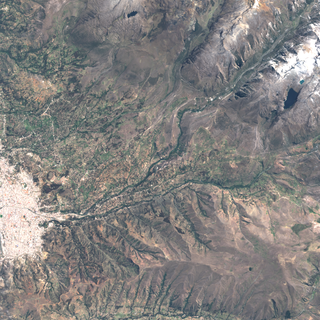}
           };
           \node[below, font=\small] at (first_image.south) {Uncropped};
    
           \node[anchor=south west,inner sep=0] (second_image) at (\spacing,0) {
               \includegraphics[width=\imagewidth\textwidth]{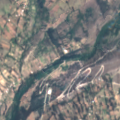}
           };
           \node[below, font=\footnotesize, align=center] at (second_image.south) {Original \\ 16~bpppb \\ $\infty$~dB};
           
           \begin{scope}[x={(first_image.south east)},y={(first_image.north west)}]
               \draw[red, thick] ($(.5,.5)-(\cropwidth/2,\cropwidth/2)$) rectangle +(\cropwidth,\cropwidth);
           \end{scope}

           \begin{scope}[x={(second_image.south east)},y={(second_image.north west)}]
               \draw[red, thick] (0,1) rectangle (1,0);
           \end{scope}
           
           \foreach \idx/\quality/\bpppb/\psnr [count=\i from 0] in {
               Best/Best/5.406/72.910,
               Very high/Very\_high/0.838/55.411,
               High/High/0.328/51.099,
               Medium/Medium/0.129/47.527,
               Low/Low/0.028/42.976,
               Very low/Very\_low/0.004/38.308
           } {
               \pgfmathsetmacro{\xpos}{(\imagewidth*\spacingmultiplier)*(\i+1)}
               \pgfmathsetmacro{\labelpos}{\xpos+0.065}
               \node[anchor=south west,inner sep=0] at (\xpos\textwidth,0) {
                   \includegraphics[width=\imagewidth\textwidth]{custom_img_libx265_12_..cubos_juliocubo1_pickle_all_\quality_1_comp_crop_center.png}
               };
               \node[below, font=\footnotesize, align=center] at (\labelpos\textwidth,0) {\quality \\ \bpppb~bpppb \\ \psnr~dB};
           }
       \end{tikzpicture}
   
       \centering
       \begin{minipage}{0.33\textwidth}
           \includegraphics[width=\textwidth]{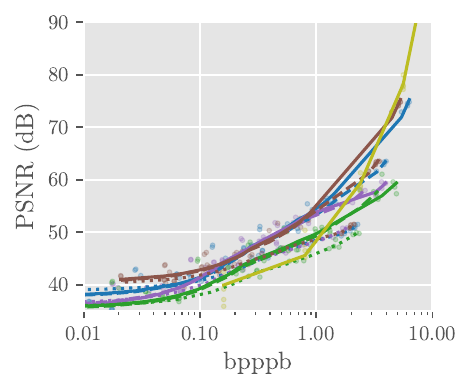}
       \end{minipage}%
       \begin{minipage}{0.33\textwidth}
           \includegraphics[width=\textwidth]{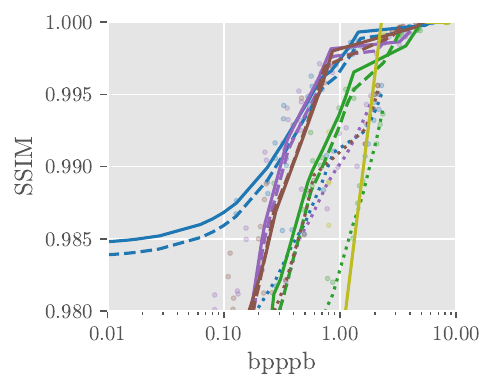}
       \end{minipage}%
       \begin{minipage}{0.33\textwidth}
           \includegraphics[width=\textwidth]{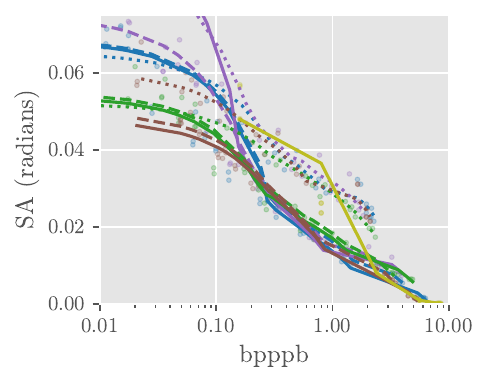}
       \end{minipage}
       \vspace{2mm}
       \includegraphics[width=1.\textwidth]{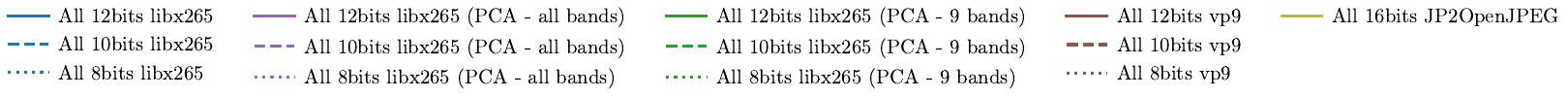}
       \caption{
           Top: Sample image from the SimpleS2 dataset showing RGB bands (original is a 16-bit integer in range 0-65535) after scaling by a factor of 1/2000 and clipping between 0 and 1 for better visualization; from left to right: original, compressed versions using \code{x265} 12 bits at different quality settings. Bottom: Compression performance metrics for the SimpleS2 dataset across three quality metrics: (a) Peak Signal-to-Noise Ratio (PSNR), (b) Structural Similarity Index (SSIM), and (c) Spectral Angle (SA) plotted against the achieved compression rate in bits per pixel per band (bpppb), for a variety of codecs (x265, vp9 and JPEG2000), at different bit depths (8, 10, 12, or 16). Higher values are better for PSNR and SSIM, while lower values are better for SA. For test ``libx265 (PCA - 9 bands)", PCA was applied over the ten bands before compression, and the least important principal component (PC) was discarded so that all nine remaining PCs could fit in just three videos of three channels each. For test ``libx265 (PCA - all bands)", PCA was applied over the ten bands before compression, and PCs were compressed differently according to their importance, such that less important PCs were still saved, albeit at a lower quality (e.g., first video would be saved at quality preset ``Best", second at ``Very high'', etc.)}
       \label{fig:custom}
  \end{figure*}
  
  We used the SimpleS2 dataset to evaluate using PCA before compressing, such that less important PCs can be compressed more aggressively or directly discarded, as in Kosheleva et al. \cite{kosheleva2003assessment}. In test ``libx265 (PCA - 9 bands)", the least important PC was discarded so that all nine remaining PCs could fit in just three videos of three channels each, instead of four for the original ten bands. For test ``libx265 (PCA - all bands)", PCs were compressed differently according to their importance, such that less important PCs were still saved, albeit at a lower quality (e.g., first video would be saved at quality preset ``Best", second at ``Very high'', etc.). Overall, the plots at the bottom of Figure~\ref{fig:custom} show that the use of PCA was negligible or even detrimental to the compression quality. Perhaps PCs are harder to encode for video codecs as they deviate farther from standard RGB data. Also, compression and decompression times were much higher for the ``libx265 (PCA - 9 bands)" test, though very similar for the ``libx265 (PCA - all bands)", as can be seen in 
  Table~\ref{tab:custom}. In terms of visual perception, images at the top of Figure~\ref{fig:custom} (no PCA used) show similar quality to those of the previous DeepExtremeCubes dataset, yet at 2x better compression ratio, likely due to the absence of clouds and other artifacts.
  
  \subsection{ERA5 --- Reanalysis weather data}
  
  \begin{table*}[!tb]
      \centering
      \caption{Comparison of different compression configurations for the ERA5 dataset. bpppb: bits per pixel per band, PSNR: Peak Signal-to-Noise Ratio, $t_{c}$: compression time (in seconds), $t_{d}$: decompression time (in seconds)}
      \makebox[\textwidth][c]{ 
      \begin{tabular}{llrrrrrrrrrrrrr}
  \toprule
  & & \multicolumn{4}{c}{Relative humidity} & \multicolumn{4}{c}{Temperature} & \multicolumn{4}{c}{Wind speed} \\
  Test & Bits & Quality & bpppb & PSNR & $t_{c}$ & $t_{d}$ & bpppb & PSNR & $t_{c}$ & $t_{d}$ & bpppb & PSNR & $t_{c}$ & $t_{d}$\\
  \midrule
  \multirow{6}{*}{\shortstack{JP2\\Open\\JPEG}} & \multirow{6}{*}{16} & Best & 8.080 & 104.069 & 143.583 & 80.693 & 5.077 & 108.793 & 125.086 & 66.239 & 6.275 & 104.749 & 131.012 & 71.915 \\
  & & Very high & 4.000 & 80.526 & 218.440 & 60.158 & 4.000 & 103.980 & 186.581 & 59.883 & 4.000 & 93.921 & 205.490 & 61.161 \\
  & & High & 0.800 & 53.330 & 185.727 & 39.411 & 0.800 & 76.343 & 156.132 & 39.847 & 0.800 & 64.941 & 171.260 & 39.804 \\
  & & Medium & 0.160 & 39.397 & 157.279 & 34.243 & 0.160 & 59.977 & 132.033 & 34.894 & 0.160 & 48.796 & 148.020 & 34.812 \\
  & & Low & 0.040 & 30.936 & 146.171 & 34.282 & 0.040 & 47.656 & 120.497 & 34.665 & 0.040 & 36.786 & 135.342 & 34.293 \\
  & & Very low & 0.008 & 22.232 & 146.185 & 32.247 & 0.008 & 36.286 & 127.730 & 31.701 & 0.008 & 27.406 & 135.717 & 30.614 \\
  \midrule
  \multirow{6}{*}{\shortstack{lib\\x265}} & \multirow{6}{*}{12} & Best & 7.320 & 74.725 & 368.723 & 48.189 & 4.944 & 81.055 & 363.321 & 38.378 & 6.409 & 76.812 & 370.582 & 44.249 \\
  & & Very high & 1.522 & 57.589 & 287.292 & 27.415 & 0.393 & 67.309 & 183.979 & 15.765 & 0.762 & 62.017 & 228.875 & 20.804 \\
  & & High & 0.833 & 52.292 & 242.304 & 21.977 & 0.210 & 63.344 & 158.872 & 13.959 & 0.431 & 57.491 & 192.631 & 17.424 \\
  & & Medium & 0.501 & 47.961 & 199.168 & 18.227 & 0.117 & 59.328 & 140.560 & 12.621 & 0.258 & 53.243 & 166.877 & 15.091 \\
  & & Low & 0.217 & 42.017 & 151.768 & 14.266 & 0.041 & 53.994 & 109.353 & 10.599 & 0.105 & 47.390 & 137.225 & 12.329 \\
  & & Very low & 0.064 & 34.932 & 106.819 & 10.742 & 0.009 & 47.700 & 68.265 & 9.143 & 0.026 & 40.417 & 95.491 & 10.178 \\
  \bottomrule
  \end{tabular}%
     }
      \label{tab:era5}
  \end{table*}
   
  \begin{figure*}[!t]
       \centering
       \begin{tikzpicture}
           \def\firstimagewidth{0.21}
           \def\imagewidth{0.105}  
           \def\spacing{0.04}     
           \def\spacingmultiplier{1.04}     
           \def\cropwidth{0.12} 
           \def\cropheight{0.16} 
           
           \node[anchor=south west,inner sep=0] (first_image) at (-\imagewidth\textwidth,0) {
               \includegraphics[angle=90,width=\firstimagewidth\textwidth]{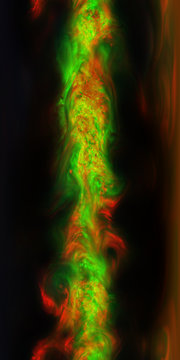}
           };
           \node[below, font=\small] at (first_image.south) {Uncropped (rel. humidity)};
    
           \node[anchor=south west,inner sep=0] (second_image) at ($(\spacing,0)+(\imagewidth\textwidth,0)$) {
               \includegraphics[angle=90,width=\imagewidth\textwidth]{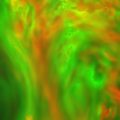}
           };
           \node[below, font=\footnotesize, align=center] at (second_image.south) {Original \\ 32~bpppb \\ $\infty$~dB};
           
           \begin{scope}[x={(first_image.south east)},y={(first_image.north west)}]
               \draw[red, thick] ($(.5,.5)-(\cropwidth,\cropheight/2)$) rectangle ($(.5,.5)+(\cropwidth*2,0.11)$);
           \end{scope}
           
           \begin{scope}[x={(second_image.south east)},y={(second_image.north west)}]
               \draw[red, thick] (0,1) rectangle (1,0);
           \end{scope}
           
           \foreach \idx/\quality/\bpppb/\psnr [count=\i from 0] in {
               Best/Best/7.320/74.725,
               Very high/Very\_high/1.522/57.589,
               High/High/0.833/52.292,
               Medium/Medium/0.501/47.961,
               Low/Low/0.217/42.017,
               Very low/Very\_low/0.064/34.932
           } {
               \pgfmathsetmacro{\xpos}{(\imagewidth*\spacingmultiplier)*(\i+1)}
               \pgfmathsetmacro{\labelpos}{\xpos+0.055}
               \node[anchor=south west,inner sep=0] at ($(\xpos\textwidth+\firstimagewidth\textwidth-\imagewidth\textwidth,0)$) {
                   \includegraphics[angle=90,width=\imagewidth\textwidth]{era5_img_libx265_12_ERA5_0_relative_humidity_\quality_40_comp_crop_center.png}
               };
               \node[below, font=\footnotesize, align=center]  at ($(\labelpos\textwidth+\firstimagewidth\textwidth-\imagewidth\textwidth,0)$)  {\quality \\ \bpppb~bpppb \\ \psnr~dB};
           }
       \end{tikzpicture}
   
       \begin{minipage}{0.33\textwidth}
           \includegraphics[width=\textwidth]{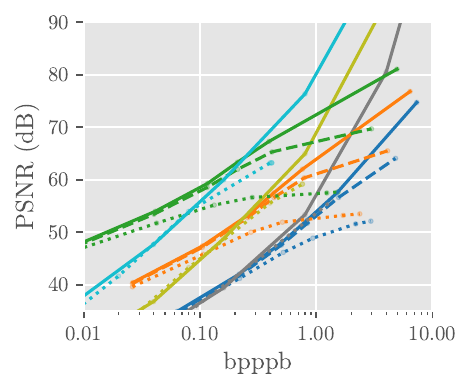}
       \end{minipage}%
       \begin{minipage}{0.33\textwidth}
           \includegraphics[width=\textwidth]{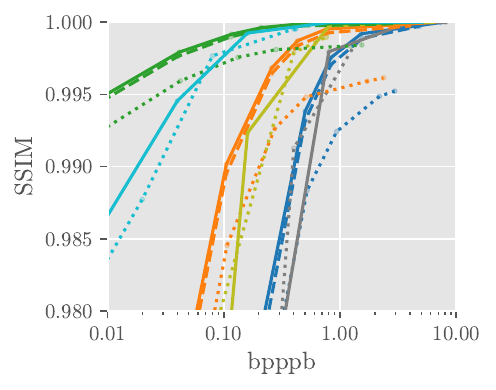}
       \end{minipage}%
       \begin{minipage}{0.33\textwidth}
           \includegraphics[width=\textwidth]{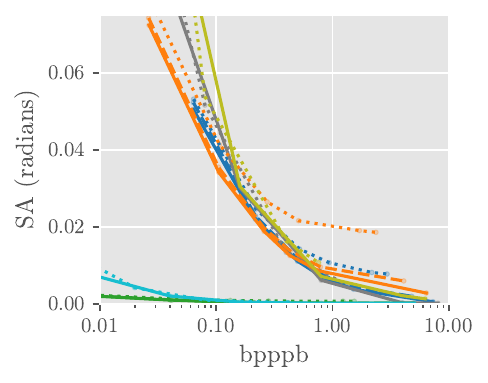}
       \end{minipage}
       \vspace{2mm}
       \includegraphics[width=1\textwidth]{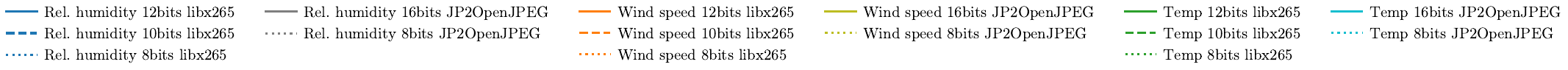}
       \caption{Top: Sample image from the ERA5 dataset showing the first three pressure levels (as RGB image channels) of the relative humidity after scaling by a factor of 2/MAX and clipping between 0 and 1 for better visualization; from left to right: original, compressed versions using \code{x265} 12 bits at different quality settings. Bottom: Compression performance metrics for the ERA5 dataset across different quality metrics: (a) Peak Signal-to-Noise Ratio (PSNR), (b) Structural Similarity Index (SSIM), and (c) Spectral Angle (SA) plotted against the achieved compression rate in bits per pixel per band (bpppb), for three variables (Relative humidity, Wind speed, and Temperature), for a variety of codecs (x265 and JPEG2000), at different bit depths (8, 10, 12, or 16). Higher values are better for PSNR and SSIM, while lower values are better for SA.}
       \label{fig:era5}
  \end{figure*}
  
  The results for the ERA5 dataset differ from the other datasets, as shown at the bottom of Figure~\ref{fig:era5}, with \code{JPEG2000} outperforming both \code{x265} and \code{vp9} at high bpppb values, and video shining at lower bpppb. As might be expected, video codecs optimized for natural RGB images do not perform as well for meteorological data. Also, some variables are much easier to compress than others, with Temperature, Wind speed, and Relative humidity achieving widely different PSNRs of approximately 80~dB, 70~dB, and 60~dB, respectively, when compressing with \code{JPEG2000} at 1~bpppb. A more detailed analysis can be found in 
  Table~\ref{tab:era5}. These experiments demonstrate the adaptability of \code{xarrayvideo} for general-purpose compression tasks within the Earth sciences. Even if not optimal for all variables within ERA5, the library could still be used to efficiently compress specific subsets of bands or variables where video encoding might offer advantages, such as Temperature.
  
  \section{Discussion}
   
  As argued, multichannel video compression has remained vastly underexplored, with Das et al. \cite{das2021hyperspectral} being the only authors exploiting spectro-spatiotemporal redundancies, though to a very limited extent.
  Unfortunately, the video they used \cite{mian2012hyperspectral} seems publicly unavailable, so a direct comparison is impossible. Still, the authors reported PSNRs of 53.18~dB using their method (and 49.08~dB using PCA + \code{JPEG2000}) at 1~bpppb, which sits below the PSNRs of 54.3, 65.9, 62.9, and 56.1 at 1~bpppb for datasets DeepExtremeCubes, DynamicEarthNet, ERA5, and SimpleS2 (as approximated with cubic interpolation on the \code{x265} data from 
  tables~\ref{tab:dynamicearthnet}-\ref{tab:era5}), hence likely being an inferior alternative.
   
  As for DL-based multichannel video compression, Han et al. \cite{han2024cra5} recently released Cra5, a 300x compressed version of the ERA5 dataset using a Variational Autoencoder Transformer, surpassing all previous techniques according to their paper. A direct comparison between their model and \code{xarrayvideo} is not straightforward, as the authors generally reported different variables and metrics than we did, always on an extremely low resolution 128$\times$256 version of the dataset, and weighting the RMSE by the latitude. Still, they reported a 0.66 Root Mean Square Error (RMSE) for variable t850 (Temperature at 850 hPa) for their validation period in 2022. Using the maximum value for t850 (308.19K) over our validation period (284 6-hour timesteps in 2022), we can approximate their PSNR at 53.38 dB. Then, using the data in 
  Table~\ref{tab:era5}, we can approximate the compression achieved using \code{x265} at that PSNR to be at 0.0368~bpppb, or an 868x compression rate for the 32~bit float data type of the uncompressed ERA5. Besides these approximate results, the Cra5 approach requires costly decoding hardware (they used an NVIDIA GPU A100-SXM4-80GB). 
   
   
  Overall, despite its advantages, lossy compression has not yet achieved global acceptance in the EO and weather communities. This is mainly because it is generally perceived that using compressed images may affect the results of posterior processing stages. However, we observe this not to be true for the two DL downstream applications (next step reflectance prediction and landcover segmentation) in which we test our compressed data, achieving negligible impact on performance metrics for models trained on the original data but evaluated on the compressed data.
  Han et al. \cite{han2024cra5} reported similar findings for their extremely compressed Cra5 dataset in the task of weather forecasting, showing even the ability of their dataset to retain the extremes. Garcia et al. \cite{garcia2010impact} go even one step further, observing that, for specific compression techniques, a higher compression ratio may lead to more accurate classification result. In summary, we argue that a dataset compressed at a sufficiently high quality is completely interchangeable with the original data for any downstream task.
   
  \section{Conclusion}
  In this work, we introduced \code{xarrayvideo}, a Python library for compressing multichannel spatiotemporal Earth system data by leveraging standard highly optimized video codecs through the widely available \code{ffmpeg} library. We evaluated the effectiveness of \code{xarrayvideo} on four representative datasets, DynamicEarthNet, DeepExtremeCubes, ERA5, and SimpleS2, demonstrating compression ratios up to 250x while maintaining high fidelity (up to 65.9dB PSNR at 1bpppb), while achieving identical performance to the original data on downstream deep learning tasks, underscoring the practical interchangeability of compressed and original data. Although recent neural compression techniques show promise, they often require specialized knowledge, hardware resources, and training, limiting their accessibility. Instead, by transforming massive datasets into compact, manageable video representations, researchers can now distribute, share, and use terabyte-scale Earth system data far more easily, as demonstrated with our 8.5Gb DynamicEarthNet and 270Gb DeepExtremeCubes releases, helping democratize the access to such resources.
  
  There are two main limitations of \code{xarrayvideo}: Firstly, despite \code{ffmpeg} being remarkably optimized and having hardware-specific (e.g., GPU) implementations widely available, loading the cubes still requires some extra processing time compared to accessing the uncompressed data. Secondly, data cubes need to be fully loaded into memory; future work could address this by allowing the loading of specific frames within the video file, hence enhancing scalability for extremely large datasets. On top of this, further lines for future research include exploring the use of the library in a broader range of use cases, including climate data, weather data, and other geospatial data sources. Besides, given the success of the approach, adapting existing codecs to accept multispectral or hyperspectral data natively would be extremely promising, as well as adding specific compression settings better suited for geophysical variables, or even developing hybrid approaches combining neural compression methods with traditional video coding techniques. By providing \code{xarrayvideo} as an open-source tool built on established standards like \code{ffmpeg}, alongside compelling use cases and datasets, we aim to democratize access to the ever-growing wealth of spatiotemporal data, fostering discoveries. Simultaneously, we hope to spark interest in lossy video compression within the EO community, hopefully leading to new and improved standards.
   
   
   
  \bibliography{citations}
  
  
  \newpage
\onecolumn

\appendices
\section{Supplementary Material}
 
 \counterwithin{table}{section} 
 \setcounter{table}{0}         
 
This Appendix contains Table~\ref{tab:codec_configs} with the configurations for the quality presets used in Tables~1-4 of the main paper.

 \begin{table}[htbp]
     \centering
          \small
     \caption{Codec configuration parameters for the different quality presets. As a single exception, for the ERA5 dataset, JP2OpenJPEG is used with qualities \code{[100, 25, 5, 1, 0.25, 0.05]}.}
     \begin{tabular}{llll}
     \toprule
     & \multicolumn{3}{c}{Codec} \\
     \cmidrule(lr){2-4}
     Quality & libx265 & VP9 & JP2OpenJPEG \\ 
     \midrule
     Best & \scriptsize\texttt{\begin{tabular}[t]{@{}l@{}}c:v: libx265\\ preset: medium\\ tune: psnr\\ crf: 51\\ x265-params:\\ \hspace{1em} lossless=1\end{tabular}} & \scriptsize\texttt{\begin{tabular}[t]{@{}l@{}}c:v: vp9\\ crf: 0\\ lossless: 1\end{tabular}} & \scriptsize\texttt{\begin{tabular}[t]{@{}l@{}}codec: JP2OpenJPEG\\ QUALITY: 100\\ REVERSIBLE: YES\\ YCBCR420: NO\end{tabular}} \\ \addlinespace[0.3em] \hline \addlinespace[0.3em] 
     Very high & \scriptsize\texttt{\begin{tabular}[t]{@{}l@{}}c:v: libx265\\ preset: medium\\ tune: psnr\\ crf: 51\\ x265-params: \\ \hspace{1em} qpmin=0:qpmax=0.01\end{tabular}} & \scriptsize\texttt{\begin{tabular}[t]{@{}l@{}}c:v: vp9\\ crf: 0\\ arnr-strength: 2\\ qmin: 0\\ qmax: 0.01\\ lag-in-frames: 25\\ arnr-maxframes: 7\end{tabular}} & \scriptsize\texttt{\begin{tabular}[t]{@{}l@{}}codec: JP2OpenJPEG\\ QUALITY: 80\\ REVERSIBLE: NO\\ YCBCR420: NO\end{tabular}} \\ \addlinespace[0.3em] \hline \addlinespace[0.3em] 
     High & \scriptsize\texttt{\begin{tabular}[t]{@{}l@{}}c:v: libx265\\ preset: medium\\ tune: psnr\\ crf: 1\end{tabular}} & \scriptsize\texttt{\begin{tabular}[t]{@{}l@{}}c:v: vp9\\ crf: 5\\ arnr-strength: 2\\ lag-in-frames: 25\\ arnr-maxframes: 7\end{tabular}} & \scriptsize\texttt{\begin{tabular}[t]{@{}l@{}}codec: JP2OpenJPEG\\ QUALITY: 35\\ REVERSIBLE: NO\\ YCBCR420: NO\end{tabular}} \\ \addlinespace[0.3em] \hline \addlinespace[0.3em] 
     Medium & \scriptsize\texttt{\begin{tabular}[t]{@{}l@{}}c:v: libx265\\ preset: medium\\ tune: psnr\\ crf: 7\end{tabular}} & \scriptsize\texttt{\begin{tabular}[t]{@{}l@{}}c:v: vp9\\ crf: 12\\ arnr-strength: 2\\ lag-in-frames: 25\\ arnr-maxframes: 7\end{tabular}} & \scriptsize\texttt{\begin{tabular}[t]{@{}l@{}}codec: JP2OpenJPEG\\ QUALITY: 15\\ REVERSIBLE: NO\\ YCBCR420: NO\end{tabular}} \\ \addlinespace[0.3em] \hline \addlinespace[0.3em] 
     Low & \scriptsize\texttt{\begin{tabular}[t]{@{}l@{}}c:v: libx265\\ preset: medium\\ tune: psnr\\ crf: 16\end{tabular}} & \scriptsize\texttt{\begin{tabular}[t]{@{}l@{}}c:v: vp9\\ crf: 20\\ arnr-strength: 2\\ lag-in-frames: 25\\ arnr-maxframes: 7\end{tabular}} & \scriptsize\texttt{\begin{tabular}[t]{@{}l@{}}codec: JP2OpenJPEG\\ QUALITY: 5\\ REVERSIBLE: NO\\ YCBCR420: NO\end{tabular}} \\ \addlinespace[0.3em] \hline \addlinespace[0.3em] 
     Very low & \scriptsize\texttt{\begin{tabular}[t]{@{}l@{}}c:v: libx265\\ preset: medium\\ tune: psnr\\ crf: 27\end{tabular}} & \scriptsize\texttt{\begin{tabular}[t]{@{}l@{}}c:v: vp9\\ crf: 30\\ arnr-strength: 2\\ lag-in-frames: 25\\ arnr-maxframes: 7\end{tabular}} & \scriptsize\texttt{\begin{tabular}[t]{@{}l@{}}codec: JP2OpenJPEG\\ QUALITY: 1\\ REVERSIBLE: NO\\ YCBCR420: NO\end{tabular}} \\ \addlinespace[0.3em]
     \bottomrule
     \end{tabular}
     \label{tab:codec_configs}
 \end{table}
  \end{document}